\documentclass[acmsmall]{acmart}

\AtBeginDocument{%
  \providecommand\BibTeX{{%
    \normalfont B\kern-0.5em{\scshape i\kern-0.25em b}\kern-0.8em\TeX}}}

\usepackage[utf8]{inputenc} 
\usepackage[T1]{fontenc}    
\usepackage{hyperref}       
\usepackage{url}            
\usepackage{booktabs}       
\usepackage{amsfonts}       
\usepackage{nicefrac}       
\usepackage{microtype}      
\usepackage{natbib}
\usepackage{comment}

\usepackage{tablefootnote}

\usepackage{amsmath, latexsym, amsthm}
\usepackage{mathtools}
\usepackage{graphicx}
\usepackage{xcolor}                 
\usepackage{hyperref}               
\usepackage{mathtext}
\usepackage{icomma}                 
\usepackage{longtable}              
\usepackage{multirow}               
\usepackage{multicol}               
\usepackage{color}
\usepackage{url}                    

\usepackage{subcaption}

\usepackage[colorinlistoftodos,bordercolor=orange,backgroundcolor=orange!20,linecolor=orange,textsize=scriptsize]{todonotes}

\newcommand{\fl}{\texttt{FL\_PyTorch}\,}

\newcommand{\RD}{\mathbb{R}^d}

\definecolor{linenumbercolor}{rgb}{0.98, 0.81, 0.69}

\definecolor{libcolor}{RGB}{152,78,163}

\definecolor{classcolor}{RGB}{0,0,255}

\definecolor{apicolor}{RGB}{255,0,0}

\definecolor{linenumbercolor}{rgb}{0.1, 0.1, 0.1}

\usepackage[ruled,vlined]{algorithm2e}
\usepackage{bm}

\newcommand{\vx}{\bm{x}}

\newcommand{\obj}{F}

\newcommand{\numClients}{\ensuremath{M}}

\newcommand{\localStep}{\tau}

\newcommand{\data}{\ensuremath{\mathcal{D}}}
\newcommand{\clientDist}{\ensuremath{\mathcal{P}}}

\newcommand{\activeClients}{\mathcal{S}}
\newcommand{\sgrad}{g}

\newcommand{\localChange}{\Delta}

\newcommand{\E}{\mathbb{E}}

\newcommand{\lr}{\eta}
\newcommand{\slr}{\lr_{s}}

\newcommand{\serveropt}{\textsc{ServerOpt}\xspace}
\newcommand{\clientopt}{\textsc{ClientOpt}\xspace}

\newcommand{\ServerGlobalState}{\textsc{ServerGlobalState}\xspace}
\newcommand{\ServerGradient}{\textsc{ServerGradient}\xspace}
\newcommand{\ClientState}{\textsc{ClientState}\xspace}
\newcommand{\LocalGradient}{\textsc{LocalGradient}\xspace}
\newcommand{\InitializeServerState}{\textsc{InitializeServerState}\xspace}

\begin{document}

\title[Optimization Research Simulator FL\_PyTorch]{\fl: Optimization Research Simulator for Federated Learning}

\author{Konstantin Burlachenko}
\email{konstantin.burlachenko@kaust.edu.sa}
\affiliation{%
  \institution{KAUST}
  \city{Thuwal}
  \country{KSA}
}

\author{Samuel Horv\'{a}th}
\email{samuel.horvath@kaust.edu.sa}
\affiliation{%
  \institution{KAUST}
  \city{Thuwal}
  \country{KSA}
}

\author{Peter Richt\'{a}rik}
\email{peter.richtarik@kaust.edu.sa}
\affiliation{%
  \institution{KAUST}
  \city{Thuwal}
  \country{KSA}
}

\renewcommand{\shortauthors}{}

\begin{abstract}
  Federated Learning (FL) has emerged as a promising technique for edge devices to collaboratively learn a shared machine learning model while keeping training data locally on the device, thereby removing the need to store and access the full data in the cloud. However, FL is difficult to implement, test and deploy in practice considering heterogeneity in common edge device settings, making it fundamentally hard for researchers to efficiently prototype and test their optimization algorithms. In this work, our aim is to alleviate this problem by introducing \fl: a suite of open-source software written in python that builds on top of one the most popular research Deep Learning (DL) framework \texttt{PyTorch}. We built \fl as a research simulator for FL to enable fast development, prototyping and experimenting with new and existing FL optimization algorithms. Our system supports abstractions that provide researchers with a sufficient level of flexibility to experiment with existing and novel approaches to advance the state-of-the-art. Furthermore, \fl is a simple to use console system, allows to run several clients simultaneously using local CPUs or GPU(s), and even remote compute devices without the need for any distributed implementation provided by the user. \fl also offers a Graphical User Interface. For new methods, researchers only provide the centralized implementation of their algorithm. To showcase the possibilities and usefulness of our system, we experiment with several well-known state-of-the-art FL algorithms and a few of the most common FL datasets.
\end{abstract}


\maketitle

\section{Introduction}
Over the past few years, the continual development of DL capabilities has revolutionised the way we interact with everyday devices. Much of this success depends on the availability of large-scale training infrastructures such as large GPU clusters and the ever-increasing demand for vast amounts of training data.
In contrary to this, users and providers are becoming increasingly aware of their privacy leakage because of this centralised data collection, leading to the creation of various privacy-preserving initiatives by industry service providers~\citep{apple} or government regulators~\citep{gdpr}.

Recently, a viable solution that has the potential to address the aforementioned issue is Federated Learning (FL). FL a term was initially proposed in \citep{mcmahan17fedavg} as an approach to solving learning tasks by a loose federation of mobile devices. However, the underlying concept of training models without data being available in a single location is applicable beyond the originally considered scenario and it turns out to be useful in other practical use-cases. For example, learning from institutional data silos such as hospitals or banks which cannot share data due to confidentiality or legal constraints, or applications in edge devices~\citep{yang2019federated, horvath2021fjord}.

The main goal of FL is to provide strong privacy protection which is obtained by storing data locally rather than transferring it to the central storage. To solve the underlying machine learning objective each client provides focused updates intended for immediate aggregation. Stronger privacy properties are possible when FL is combined with other technologies such as differential privacy~\citep{dwork2008differential} and secure multiparty computation (SMPC) protocols such as secure aggregation~\citep{bell2020secagg}. 

Recently, federated learning has seen increasing attention not only in academia but also in the industry which already uses FL in their deployed systems. For instance,
Google use it in the Gboard mobile keyboard for applications including next word prediction~\citep{hard18gboard}, emoji suggestion~\citep{gboard19emoji} or “Hey Google” Assistant~\citep{googleassistant2021}. Apple uses FL for applications like the QuickType keyboard and the vocal classifier for “Hey Siri”~\citep{apple19wwdc}. In the finance space, FL is used to detect money laundering~\citep{webank2020} or financial fraud~\citep{intel2020}. In the medical space, federated learning is used for drug discovery~\citep{melloddy2020}, for predicting COVID-19 patients’ oxygen needs~\citep{nvidia2020} or medical images analysis~\citep{owkin2020} and others.

\begin{algorithm}[t]
    \DontPrintSemicolon
    \SetKwInput{Input}{Input}
    \SetAlgoLined
    \LinesNumbered
    \Input{Initial model $\vx^{(0)}$, \textsc{ClientOpt}, \textsc{ServerOpt}}
    Initialize server state $H^{0}=\InitializeServerState()$\;
     \For{$t \in \{0,1,\dots,T-1\}$ }{
      Sample a subset $\activeClients^{(t)}$ of available clients\;
      Generate state: $s^{(t)} = \ClientState(H^{(t)})$ \;
      Broadcast $(\vx^{(t)},s^{(t)})$ to workers \;
      \For{{\it \bf client} $i \in \activeClients^{(t)}$ {\it \bf in parallel}}{
        Initialize local model $\vx_i^{(t,0)}=\vx^{(t)}$\;
        \For {$k =0,\dots,\localStep_i-1$}{
            Compute local stochastic gradient $\sgrad_i =\LocalGradient(\vx_i^{(t,k)}, s_t)$\;
            Perform local update $\vx_i^{(t,k+1)} = \textsc{ClientOpt}(\vx_i^{(t,k)}, \sgrad_i, {k}, t)$\;
        }
        Compute local model changes $\localChange_i^{(t)} = \vx_i^{(t,\localStep_i)} - \vx_i^{(t,0)}$\;
        Create local state update: $U_i^{(t)} = \textsc{LocalState}(\vx^{(t)}, \vx_i^{(t,\localStep_i)})$\;
        Send $(\localChange_i^{(t)},U_i^{(t)})$ to Master.\;
      }
      Obtain $(\localChange_i^{(t)},U_i^{(t)}), \forall i \in \activeClients^{(t)} $. \;
      Compute $G^{(t)} = \ServerGradient(\{\localChange_i^{(t)},U_i^{(t)}\}_{i \in S^{(t)}}, H^{(t)})$ \; 
      Update global model $\vx^{(t+1)} = \textsc{ServerOpt}(\vx^{(t)}, G^{(t)},\slr,t)$
      \;
      Update: $H^{(t+1)}=\ServerGlobalState(\{\localChange_i^{(t)},U_i^{(t)}\}_{i \in S^{(t)}}, H^{(t)})$ \;
     }
     \caption{Generalized Federated Averaging}
     \label{algo:generalized_fedavg}
\end{algorithm}

To enable research in federated learning, several frameworks have been proposed  including \texttt{LEAF} \citep{caldas2018leaf}, \texttt{FedML} \citep{he2020fedml}, \texttt{Flower} \citep{beutel2020flower}), (\texttt{PySyft} \citep{ryffel2018generic}, \texttt{TensorFlow-Federated} (TFF) \citep{TFF2019}, \texttt{FATE} \citep{yang2019federated}, \texttt{Clara} \citep{ClaraTraining}, \texttt{PaddleFL} \citep{ma2019paddlepaddle}, \texttt{Open FL} \citep{OpenFLFramework}. These frameworks are mainly built with a focus to be deployed on real world systems while also providing user with an ability to run experiments with the same code. This desired property often comes with the price that the entry bar for researchers to extend or experiment with these frameworks is limited in a sense that they either need to have extensive experience with distributed systems or require assistance from experts on given framework.

In our work, we decided to take one step back and focus on the construction of a framework that, although not aimed for being deployed to edge devices as primary goal, can provide a useful simulation environment for researchers with the following goals:
\begin{itemize}
    \item \textit{Low Entry Bar/ Simplicity}: We aim our tool to be as simple as possible for the user while providing all necessary functionalities. 
    \item \textit{Extensibility}: It is easy to bring your own algorithm or dataset or extend existing ones. We aim to achieve this by providing universal abstractions with a sufficient level of flexibility to experiment with existing and novel approaches to advance the state-of-the-art.
    \item \textit{Hardware Utilisation}: Cross-device FL experiments are usually of much smaller scale comparing to the centralized setting. This is mainly because of limited device capabilities. Running such experiments on GPU can lead to the under-utilisation of available hardware. We aim to resolve this via the ability to parallelise clients' computation.
    \item \textit{Easy Debugging}: Debugging multi-process or multi-thread systems is hard. We only require user to provide a single thread implementation which is automatically adjusted to multi-GPU and multi-node setup.
\end{itemize}
To the best of our knowledge, there is no such tool that could provide a sufficient level of freedom and is simple to use and therefore, we design \fl to achieve the aforementioned goal.

\fl is an optimization research simulator for FL implemented in Python based on \texttt{PyTorch}~\citep{pytorch}. \fl is a simple to use tool with its own Graphical User Interface (GUI) implemented in PyQt~\citep{pyqt_docu}. During the simulation process, the selected local CPUs/GPUs are accessed in a parallel way. In addition, there is a possibility to use remote CPUs/GPUs. Remote devices are required to have a TCP transport layer for communicating with the master. Regarding supported devices, we target server and desktop stations running on Linux, macOS, or Windows OS for efficient simulations.

For the paper organization, we introduce the general FL minimization objective in Section~\ref{sec:gen_objective}. Subsequently, we provide a deep dive into our \fl system in Section~\ref{sec:fl_pytorch} and, finally, we demonstrate \fl capabilities by concluding several experiments on multiple FL baselines and the most used FL dataset in Section~\ref{sec:exp}.

\section{FL Objective and FedAVG}
\label{sec:gen_objective}
In this section, we introduce the FL objective in its general form
\begin{align}
    \obj(\vx) = \E_{i \sim \clientDist}[ \obj_i(\vx)] \quad \text{where} \ \obj_i(\vx) = \E_{\xi \in \data_i}[f_i(\vx, \xi)] + R(x). \label{eqn:global_obj}
\end{align}

The global objective $\obj$ is an expectation over local objectives $\obj_i$ over the randomness inherited from the client distribution $\clientDist$, and the local objectives $\{\obj_i\}$ have the form of an expectation over the local datasets $\{\data_i\}$. The $\{\obj_i\}$ has an additional regularization term that is useful for incorporating prior knowledge of parameters to find $x$.

In the case of the finite number of clients and local data points, both global objectives $\obj$ and local losses $\{\obj_i\}$ can be written as simple weighted averages in the empirical risk minimization form (ERM). The most common algorithm to solve \eqref{eqn:global_obj} is federated averaging \texttt{FedAVG}~\cite{mcmahan17fedavg}. This algorithm divides the training process into communication rounds. At the beginning of the $t$-th round ($t \geq 0$), the server broadcasts the current global model $\vx^{(t)}$ to a random subset of clients $\activeClients^{(t)}$ (often uniformly sampled without replacement in simulations). Then, each sampled client performs $\localStep_i$ local SGD updates on its own local dataset and sends its local model update $\Delta_i^{(t)}=\vx_i^{(t,\localStep_i)}-\vx^{(t)}$ to the server. Finally, the server uses the aggregated model updates to obtain the new global model as follows:
\begin{align}
    \vx^{(t+1)} 
    = \vx^{(t)} + \frac{\sum_{i \in \activeClients^{(t)}} p_i \Delta_i^{(t)}}{\sum_{i \in \activeClients^{(t)}} p_i}. \label{eqn:upadte_fedavg}
\end{align}
where $p_i$ is the relative weight of client $i$. The above procedure will repeat until the algorithm converges. In the \emph{cross-silo} setting where all clients participate in the training at every round, we have $\activeClients^{(t)}=\{1,2,\dots,\numClients\}$. 

\section{FL Optimization Simulator}
\label{sec:fl_pytorch}
\fl is a system that has been built using Python programming language, and it is based on the DL framework \texttt{PyTorch}. We have made the code repository publicly available\footnote{\href{https://github.com/burlachenkok/flpytorch}{https://github.com/burlachenkok/flpytorch}}.

Its backbone consists of a general form of FedAVG displayed in Algorithm~\ref{algo:generalized_fedavg} partially inspired by Algorithm 1 in~\citep{reddi2020adaptive}. Our proposed general form preserves the standard structure of federated optimization where in each round $t$, subset $\activeClients^{(t)}$ of all available clients is selected. As a next step, the master generates its state $s_t$ which is broadcasted together with the current copy of the global solution $\vx^{(t)}$ to the selected subset of clients $\activeClients^{(t)}$. Afterwards, each participating client initializes its local solution $\vx_i^{(t,0)}$ to be a copy of the global solution $\vx^{(t)}$. Each of these clients then performs $\tau_i$ steps using the local optimizer $\clientopt$ with gradient estimated by the $\LocalGradient$ function. After this step, each client computes the model and state update, which are then sent back to the master which estimates the global update direction using the $\ServerGradient$ function. This estimate is used in $\serveropt$ that updates the global solution to its new value $\vx^{(t+1)}$. Lastly, the server global state is updated. In Section~\ref{sec:algorithms}, we show that this general scheme captures all standard algorithms, thus our scheme is sufficient and we believe that it gives researchers a sufficient level of freedom to develop and experiment with each component of our general scheme to push both practical and theoretical FL state-of-the-art. 

Our current implementation is fully determined and allows us to configure up to 51  parameters which can be specified either through our easy to use GUI tool or directly via the command line that we discuss in the next subsection. These parameters can be grouped into 4 categories based on their function: 
\begin{itemize}
\item \textit{Server Optimizer}: the number of communication rounds $T$; the number of sampled clients per round; global initial local learning rate, global learning rate schedule, and global optimizer with its parameters such as momentum, the name of the algorithm to execute.

\item \textit{Local Optimizer}: the number of local epochs or local iterations $\tau_i$; batch size for data loading; local optimizer with its parameters.

\item \textit{Model and Data}: model's and dataset's names.

\item \textit{System Setup}: directory to store run metadata, target compute devices, usage of remote compute devices, logging level, number of workers for loading dataset, random seed, thread pool sizes, experiments grouping, user defined comment, enable usage of NVIDIA Ampere GPU Tensor Cores, optional cleanup of PyTorch GPU cache at the end of each round, device for store client state\footnote{For some optimization algorithms with a client state and with partial participation, it may be better to temporarily store the client state in CPU DRAM memory instead of GPU memory.}.

\end{itemize}

\subsection{Frontend}
As we mention previously, \fl supports two modes for running federated optimization--Graphical User Interface and Command Line Interface. Below, we discuss both of these in detail.

\subsubsection{Graphical User Interface (GUI)}
We implemented our Graphical User Interface using the PyQt5 GUI framework~\citep{pyqt_docu}. This GUI framework supports all the standard desktop operating systems such as  Windows, Linux, or macOS. In addition, the GUI part of \fl has built-in VNC server support. If the target machine has not a native windows manager system one can use this mode and connect to GUI part via VNC Viewer software.

\begin{figure*}[t]
\begin{tabular}{cc}
  \includegraphics[width=67mm]{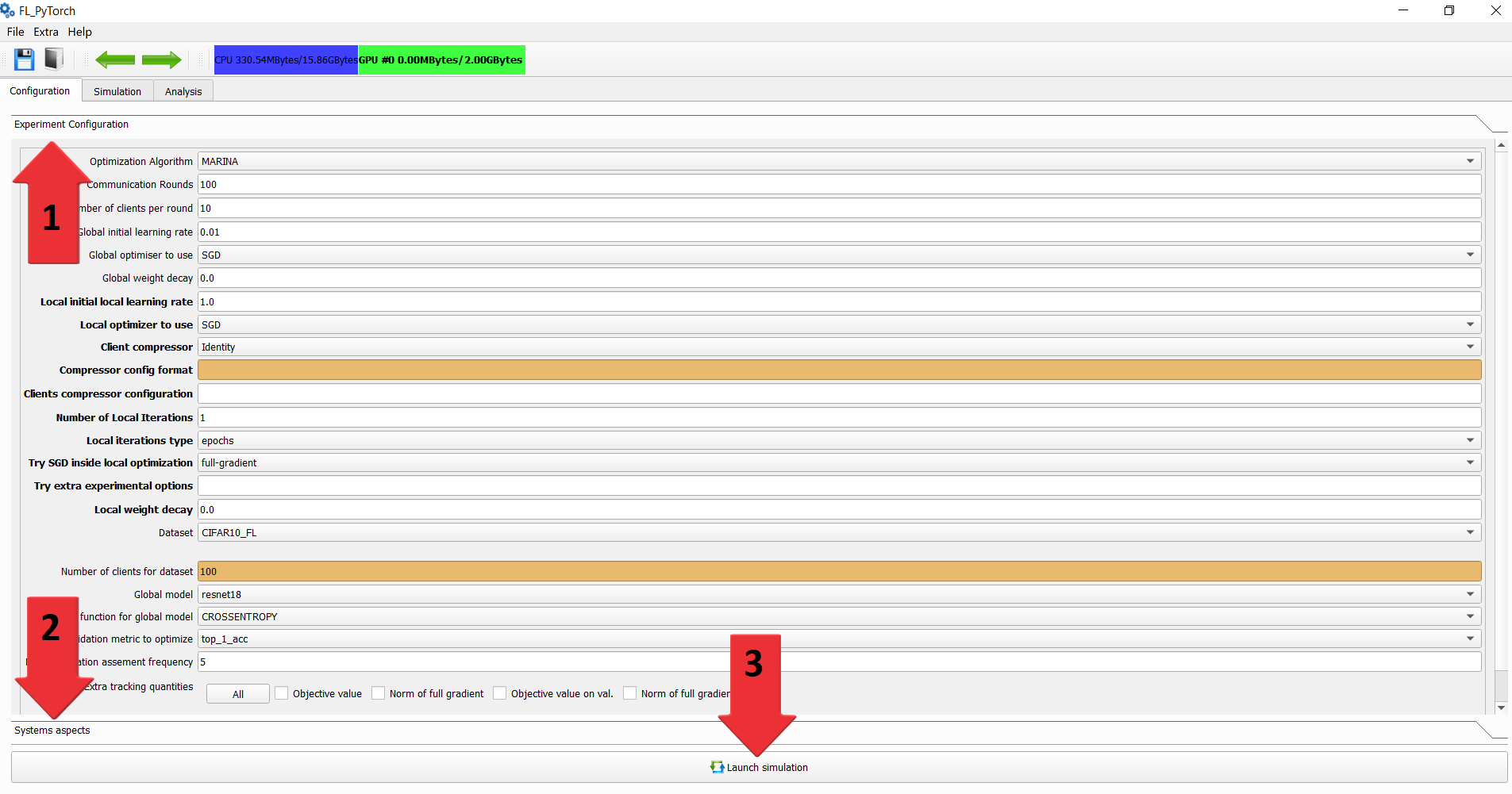} &
  \includegraphics[width=67mm]{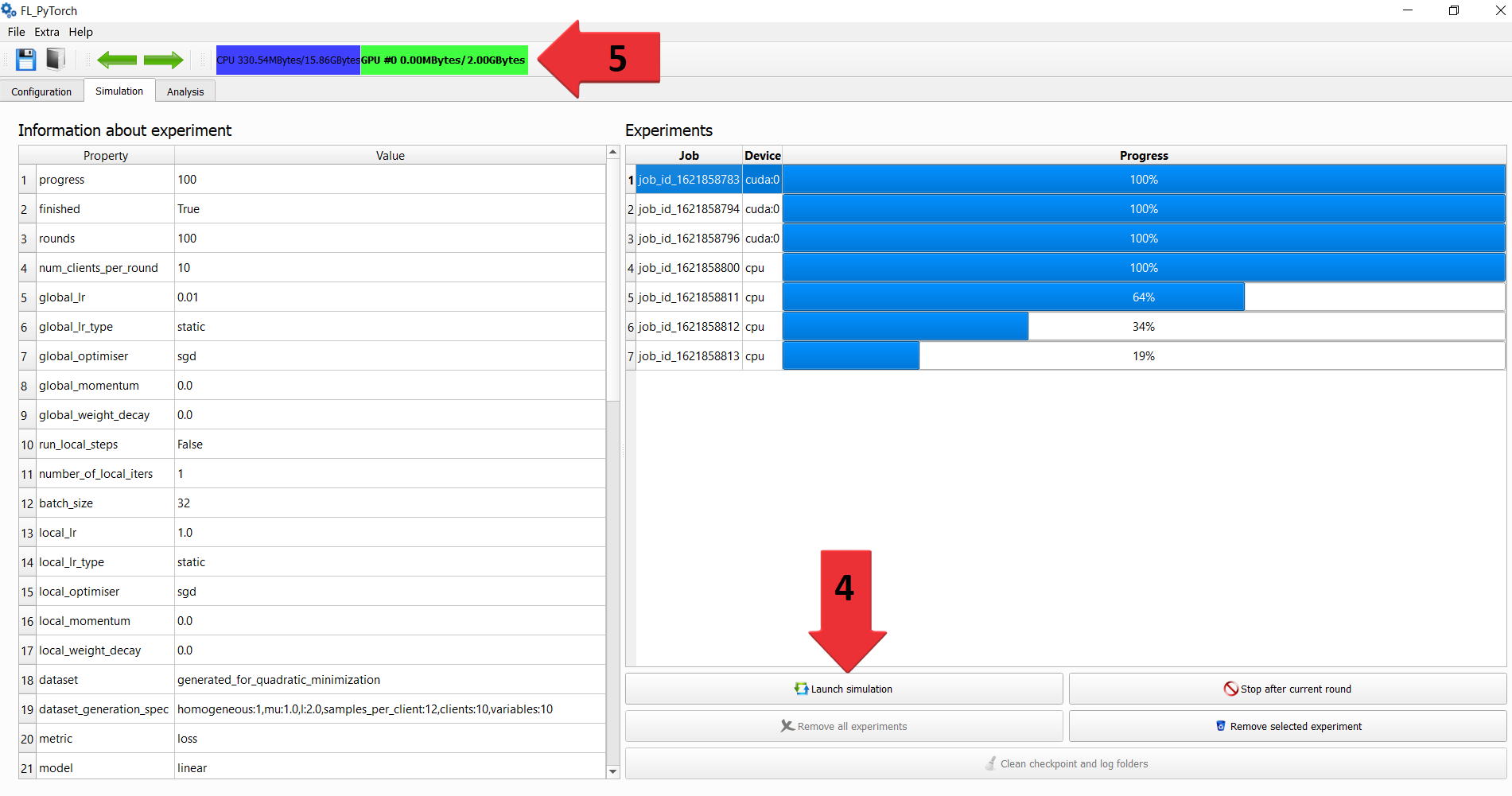} \\
(a) The main tab of the \fl's GUI. & 
(b) The Progress tab of the \fl's GUI. \\[2pt]

 \includegraphics[width=67mm]{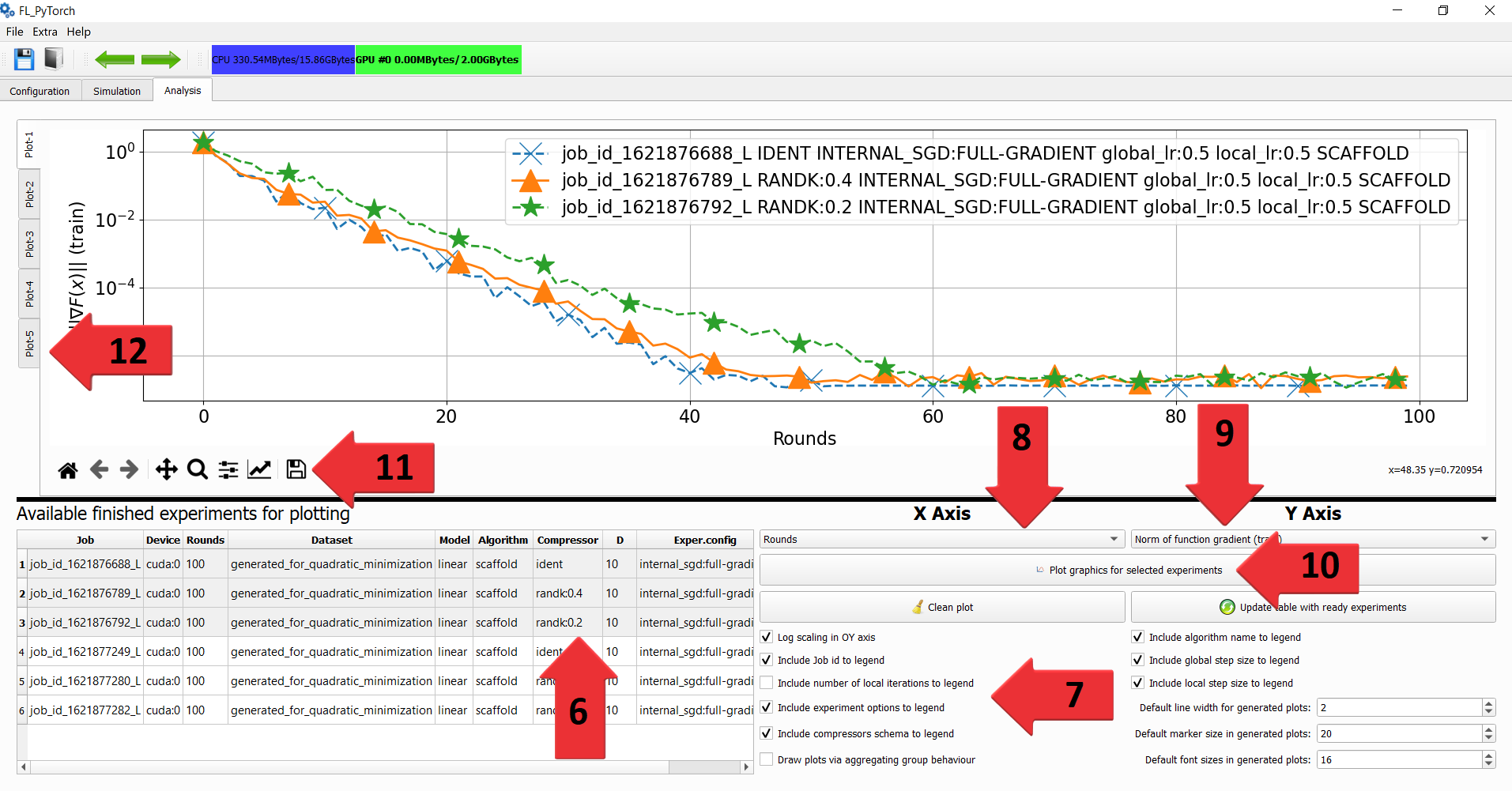} &
 \includegraphics[width=67mm]{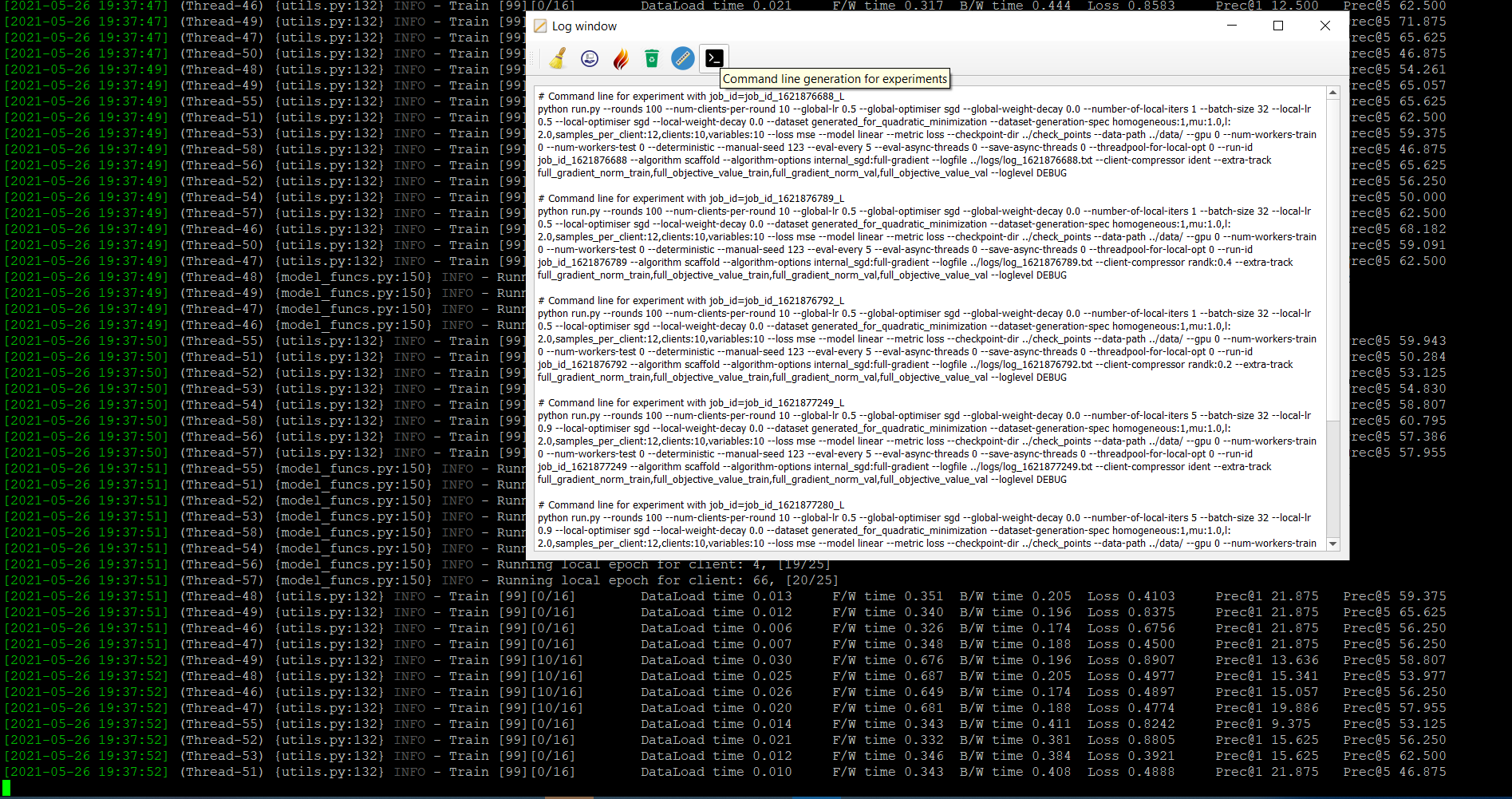} \\
(c) The Analysis tab of the \fl's GUI.  &
(d) Console output during simulation and GUI log. \\[2pt]
\end{tabular}
\caption{Graphical User Interface of the simulator}
\label{fig:fl_gui_simulation_gui}
\end{figure*}

To get a better picture of our GUI, we display its main menu provided in Figure~\ref{fig:fl_gui_simulation_gui}, (a), (b) and (c) are the 3 main components of our GUI and they help users to build their run setup with the steps depicted as numbered red arrows. 1.) experiment configuration,  2.) system setup configuration, 3.) thread pool setup, 4.) button to launch experiments,  5.) navigation pane, 6.) system monitoring, 7.) plotting setup -- experiments selection, 8.) plotting setup -- format 8.) plotting setup -- x-axis, 9.) plotting setup -- y-axis, 10.)   plotting setup -- generate plots, 11.) plotting setup --  save option, 12.)  plotting setup -- clean output option.

For purpose of obtaining information regarding the currently used system and environment, the user can select in GUI Help$\to$About.

Some extra configurable displayed information from GUI includes a number of currently executing experiments, showing the available size of physical DRAM and swap memory in OS.

\subsubsection{Command Line Interface (CLI)}
Another alternative option to pass the run arguments is directly through the console via Command Line Interface (CLI). 
We also provide assistance to see the available arguments' options that can be accessed by launching \texttt{./run.sh} script provided in the supplementary materials with the\texttt{--help} option. 
To enable an efficient way to switch between GUI and CLI, we provide an option to recover arguments passed via the GUI tool in the command line format, displayed in Figure~\ref{fig:fl_gui_simulation_gui}. It can be used to generate a command line for already finished experiments which is currently configured in GUI, but not yet launched. (d). The CLI interface is used to instantiate remote workers in remote machines, for providing extra remote compute resources for the simulation process. Also, there is an integration with WandB online plotting tool. \footnote{\href{https://wandb.ai/home}{https://wandb.ai/home}}. Using it from CLI will allow you to monitor the progress of experiments. It's possible to use that tool from CLI and GUI. To perform an early stopping of the experiment, you should send Posix SIGTERM or SIGINT signals to it. In that case, the experiment will be terminated, but simulation results will be appropriately saved.

\subsection{Optimization Algorithms}
\label{sec:algorithms}
As discussed previously, our general level of abstraction introduced in Algorithm~\ref{algo:generalized_fedavg} allows for a sufficient level of freedom to implement standard and also more exotic FL optimization algorithms. In the current version, we implemented following state-of-the-arts methods considered in the literature: Distributed Compressed Gradient Descent (\texttt{DCGD})~\citep{alistarh2017qsgd,khirirat2018distributed,horvath2019natural} \texttt{FedAVG}~\citep{mcmahan17fedavg}, \texttt{SCAFFOLD}~\citep{karimireddy2020scaffold}, \texttt{FedProx}~\citep{li2018federated}, \texttt{DIANA}~\citep{diana}, \texttt{MARINA}~\citep{gorbunov2021marina},\texttt{PP-MARINA}~\citep{gorbunov2021marina}, \texttt{EF21}~\citep{richtarik2021ef21},\texttt{EF21-PP}~\citep{fatkhullin2021ef21}, \texttt{COFIG}~\citep{zhao2021faster} and \texttt{FRECON}~\citep{zhao2021faster}.

Some optimization algorithms require a client to store in its internal state-specific information about algorithm-specific shifts that have a notion about gradient estimator. The analysis is very often irrelevant for the specific strategy of initializing shifts, but in practice, different initialization policies may have consequences in convergence speed. We provide two policies with initializing shift by zero or by full gradient at $x^{(0)}$.

As an example, we show how Algorithm~\ref{algo:generalized_fedavg} is adjusted to \texttt{FedAVG} and \texttt{SCAFFOLD}. For \texttt{FedAVG}, both global state is an empty dictionary. Locally, we run $\tau_i$ iteration, which is usually set to be a constant $T$ for the theoretical and size of the local dataset for the experiments. $\LocalGradient$ returns the unbiased stochastic gradient at $\vx^{(t,k)}$ and $\clientopt$ is Stochastic Gradient Descent (\texttt{SGD}) with given step-size $\eta_l$. Similarly to the global state, local step update is none. $\ServerGradient$ is a (weighted) average of local updates and $\serveropt$ is \texttt{SGD} with fixed step-size $1$. For the \texttt{SCAFFOLD}, there are $2$ changes comparing to \texttt{FedAVG}. Firstly, the global state $s^{(t)}$ is a non-empty vector of the same dimension as $\vx^{(t)}$ initialized as zero and each $\clientopt$ has its own local  $s_i^{(t)}$, which is set to be a stochastic local gradient at $\vx^{(t)}$. $\LocalGradient$ then returns the unbiased stochastic gradient at $\vx^{(t,k)}$ shifted by $s_i^{(t)} - s_i^{(t)}$. Secondly, the local model update is sent to master together with the local state update, which is set to the difference between current and previous communication round $U_i^{(t)} = s_i^{(t)} - s_i^{(t-1)}$. The $\ServerGlobalState$ is then updated using the average of $U_i^{(t)}$'s.

\subsection{Supported Compressors}
In federated learning (especially in the cross-device setting), clients may experience severe network latency and bandwidth limitations. Therefore, practical FL algorithms generally use a communication reduction mechanism to speed up the training procedure. Three common methods to reduce the communication cost are:
\begin{enumerate}
    \item to reduce the communication frequency by allowing local updates;
    \item to reduce communication traffic at the server by limiting the participating clients per round;
    \item to reduce communication volume by compressing messages.
\end{enumerate} 
We naturally support the first and the second option and we also added support for the compression. \fl allows compressing messages both from the server to local clients and vice-versa. We support several unbiased and biased compressors. The supported unbiased compressor are: Identical compressor (no compression), Lazy or Bernoulli compressor (update is communicated with probability $p$ rescaled for the update to be unbiased), Rand-$K$ (only random $K$ coordinates are preserved uniformly at random rescaled for the update to be unbiased), Natural compressor~\citep{horvath2019natural}, Standard dithering~\citep{alistarh2017qsgd}, Natural Dithering~\citep{horvath2019natural}, TernGrad~\citep{wen2017terngrad} and QSGD compressor~\citep{alistarh2017qsgd}. The supported biased compressor are: Top-$k$~\citep{beznosikov2020biased} and Rank-$k$~\citep{safaryan2021fednl}. For Top-$k$ and Rank-$k$ compressors, parameter $K$ can be specified as a percentage in $[0,100]$ from $d$ or as an absolute number greater than $1$.

Also, we provide means to construct new compressors via function composition and probabilistic switching from existing.

\begin{figure*}[t]
\begin{tabular}{cc}
  \includegraphics[width=67mm]{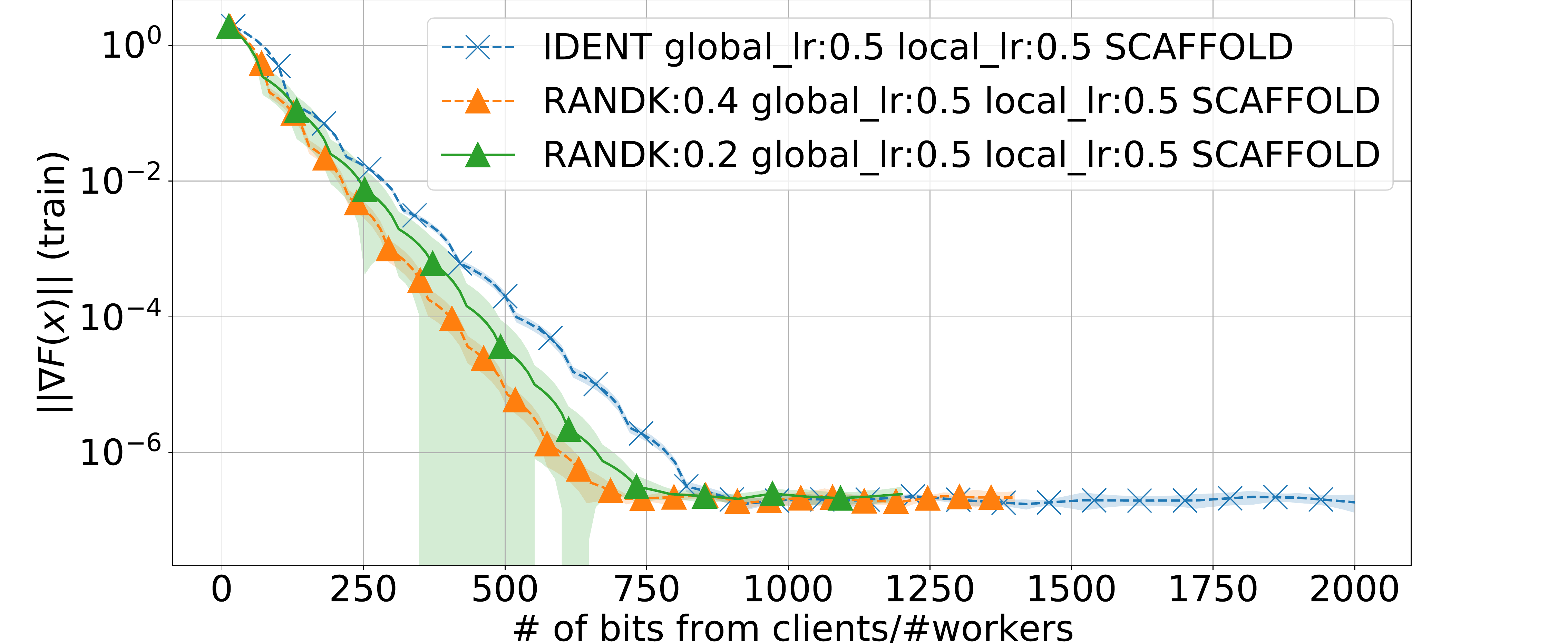} &
  \includegraphics[width=67mm]{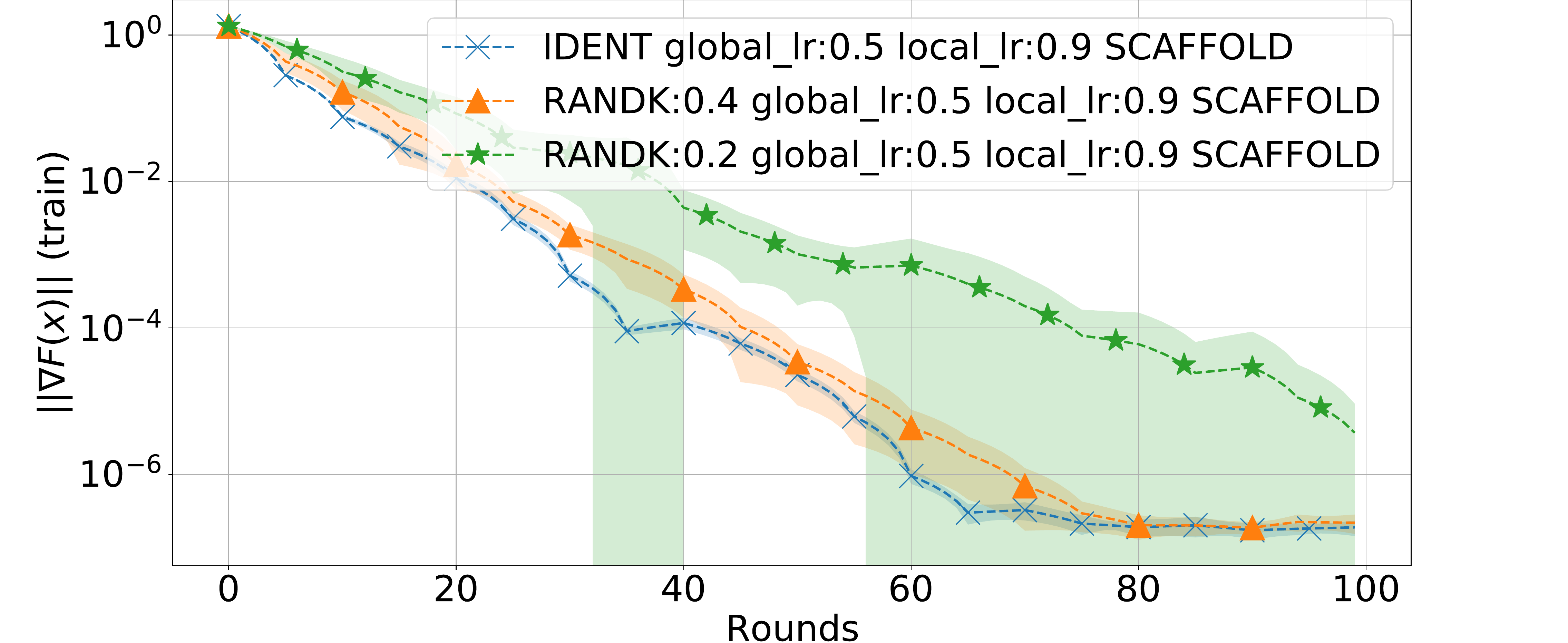} \\
(a) One local iteration, convergence in gradient. & 
(b) Five local iteration,convergence in gradient. \\[2pt]
 \includegraphics[width=67mm]{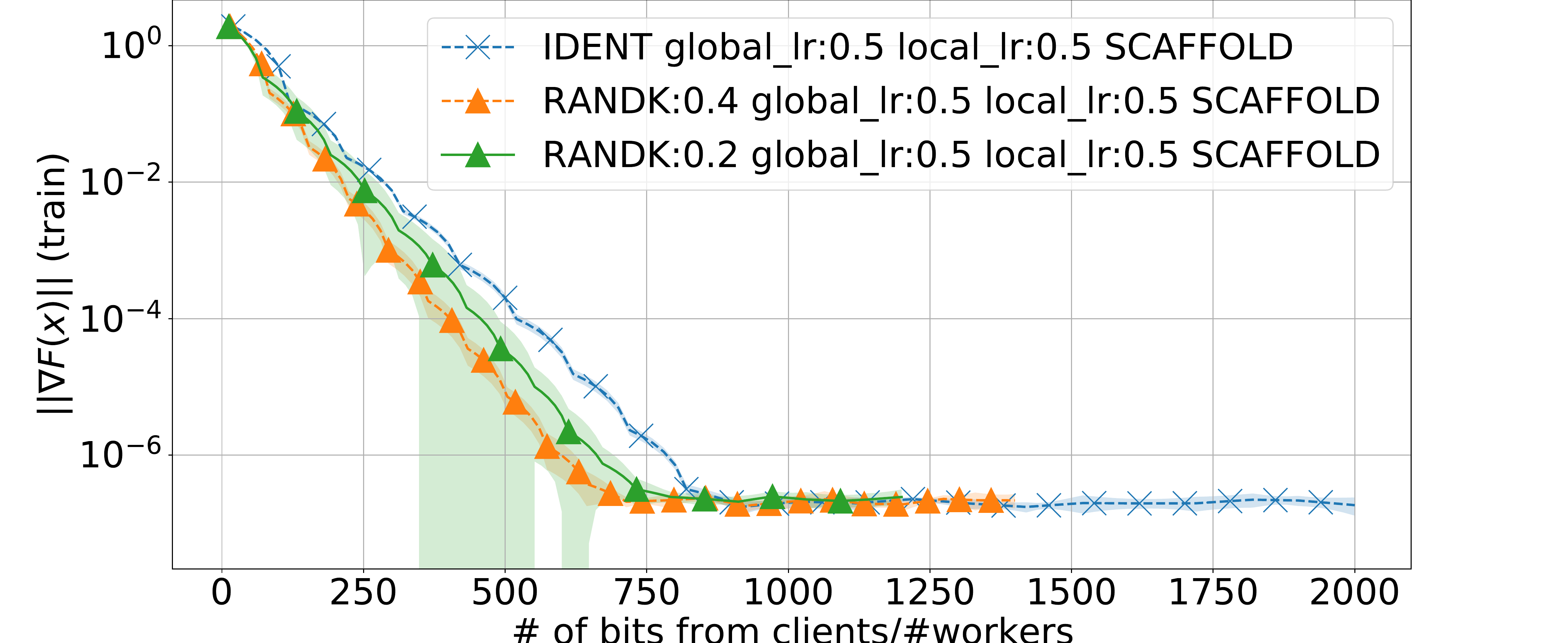} &
 \includegraphics[width=67mm]{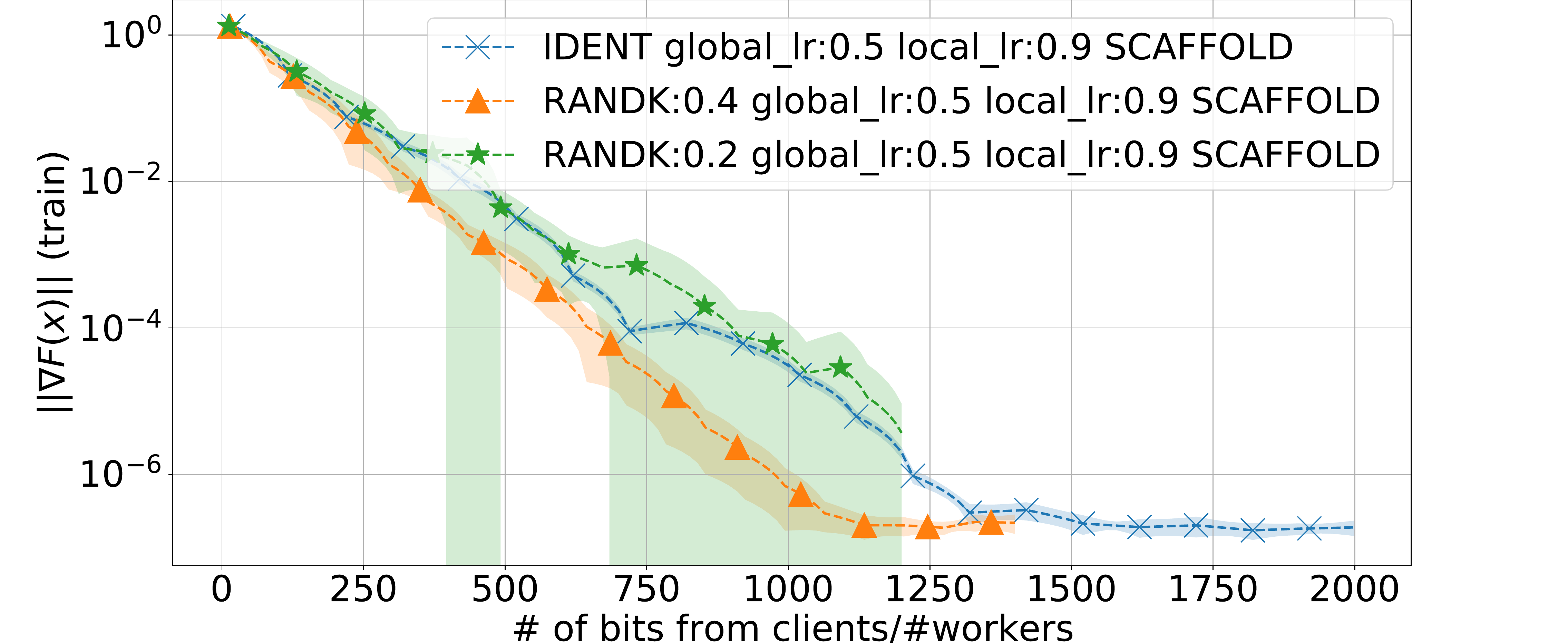} \\
(c) One local iteration, average communication.  &
(d) Five local iterations, average communication. \\[2pt]
\end{tabular}
\caption{Function gradient diminishing for 1 and 5 local iterations of SCAFFOLD for quadratic minimization. Mean and variance have been estimated across 10 realizations.}
\label{fig:scaffold_experiment_convex}
\end{figure*}

\subsection{Supported models(or pattern) structures}

Current \fl's implementation allows users to experiment with the following list of image classification models: 
 ResNets~\citep{resnet} (18, 34, 50, 101, 152), VGGs~\citep{vgg} (11, 13, 16, 19) and WideResNets~\citep{wideresnet} (28\_2, 28\_4, 28\_8), Logistic Regression.
 
Besides DL models, we provide support for simple quadratic loss function to enable users to explore algorithms in this simplistic regime or debug their implementation. 
 We synthetically generate our objective such that local loss is of the form
\begin{equation}
\label{eq:synthetic_quadratic}
\min_{x \in \RD} \dfrac{1}{n_i} \sum_{i=1}^{n_i}  \| a_i^\top x - b_i\|^2.
\end{equation}
Users are able to specify: dimensionality $d$, strong convexity parameter $\mu$ and $L$ smoothness constant for \eqref{eq:synthetic_quadratic} and the number of samples per each client. In addition, we consider two settings wherein the first case all clients optimizes the same objective, i.e., iid setting. In the second scenario, each client has a different objective to optimise, i.e.,  non-iid setting.

\subsection{Tracking metrics and experiments analysis}
An important feature to evaluate experiments is the ability to track all important quantities during the run of the algorithms. \fl allows users to track the number of communication rounds, loss, accuracy, the norm of computed gradient, the norm of the full objective gradient, function values, number of gradient oracle calls, used GPU memory, number of bits which would be sent from workers to master in an actual real-world system, number of clients per round, wall-clock execution time. Furthermore, we provide a dedicated visualization tab in our GUI tool which allows user to load and visualize their experiments in an interactive fashion in several plots. You can observe such quantities during experiment execution via two-dimensional plots.

Each experiment has a lot of associated information. We provide means to configure the view of experimental results via selecting needed attributes, sorting experiments by some attribute in the analysis tab, and reordering experiments manually.

\subsection{The internals of the system}
To reach one of our goals of making \fl computationally effective, we analyzed various aspects of the \texttt{PyTorch} infrastructure, including the place of \texttt{PyTorch} at NVIDIA computation stack, \texttt{PyTorch} initialization, its forward and backward computation, and speed of typical buses used in a local node setup. This analysis led us to the system design, which asynchronously exploits available hardware. 

The Algorithm~\ref{algo:generalized_fedavg} has several independent parts, and instead of sequential execution of these parts in a simulated environment, independent parts can be partition across independent thread-pools of workers. In our implementation, each worker within a thread pool is a separate CPU thread that lives within a Python interpreter process launched in the operation system (OS) if the system's resources allow it. Since \texttt{PyTorch} and therefore \fl allow launching of the computations in a GPUs, for providing separate CPU threads ability to work independently, each thread has its separate GPU CUDA stream to submit its computation work for GPU independently. By design, each worker with the same purpose is assigned to its task-specific thread pool with a thread-safe tasks queue. There are three specific types of tasks -- deferred execution, process request for finish execution, and wait on the completion of all submitted work.

Finally, we stress that all this is happening "behind the curtain," and user interaction, including implementation of the new methods, is agnostic to the aforementioned implementation details.
\subsection{Bringing custom Algorithm/ Model/ Data}
\label{sec:custom_alg}

As described in Algorithm~\ref{algo:generalized_fedavg}, each algorithm supported by \fl \\ requires implementation of \InitializeServerState,  \ClientState, \LocalGradient, \textsc{ClientOpt},   \textsc{LocalState},   \textsc{ServerOpt}, \ServerGradient, \ServerGlobalState. These are standard centralized-like \texttt{PyTorch} functions that need to be provided to \fl in order to run simulations. 

One example of adding new algorithm might be \texttt{SCAFFOLD} assuming that we are given implementation of \texttt{FedAvg} in the required Algorithm~\ref{algo:generalized_fedavg} format. As described in Section~\ref{sec:algorithms}, one needs to set \ClientState to return global shift as proposed in \citep{karimireddy2020scaffold} and update \LocalGradient to account for both global and local shifts. One further needs to define \textsc{LocalState} to return local shift update, see \citep{karimireddy2020scaffold}  for the detailed formula.  The last modification is in \ServerGlobalState, which updates global shift using the local shift updates. 

For the datasets, we require users to provide it in a form that is loadable via \texttt{PyTorch}'s \texttt{DataLoader}s and for the models, the required format is  \texttt{PyTorch}'s \texttt{nn.Module}. 

\section{Experiments}
\label{sec:exp}
In this section, we provide several experiments to showcase potential use cases of \fl. In general, \fl can serve as a simulator for researchers to test and compare different algorithms, extend them and analyze each algorithmic component in a plug-and-play fashion as described in Section~\ref{sec:fl_pytorch} and Algorithm~\ref{algo:generalized_fedavg}. 

For instance, one might be interested in how to reduce the communication burden of a given algorithm. In the literature, three approaches have been proposed. For the first approach, one performs local updates to reduce the frequency of the global model updates, thus also reducing communication frequency. By this strategy,  communication cost per client model update can be effectively reduced by a number of local steps. However, while often in practice this leads to improvement, it is not always guaranteed, as it was theoretically shown that there is a trade-off between communication reduction and convergence~\citep{charles2021convergence}. Another possibility is to employ compression techniques that reduce the bits transmitted between the clients and the server. This compression operator should be selected carefully as it comes with an extra variance and it might affect the convergence. The third option is to employ partial participation for which in each round only sampled subset of clients participates. There is also a trade-off for this strategy as although having only a few clients in each round has a positive effect on reducing communication, it might negatively affect the quality of the obtained solution due to inexact updates based only on sub-sampling. \fl supports experimentation with all 3 approaches and below, we include 3 such examples. 

\begin{figure*}[t]
\begin{tabular}{cc}
  \includegraphics[width=67mm]{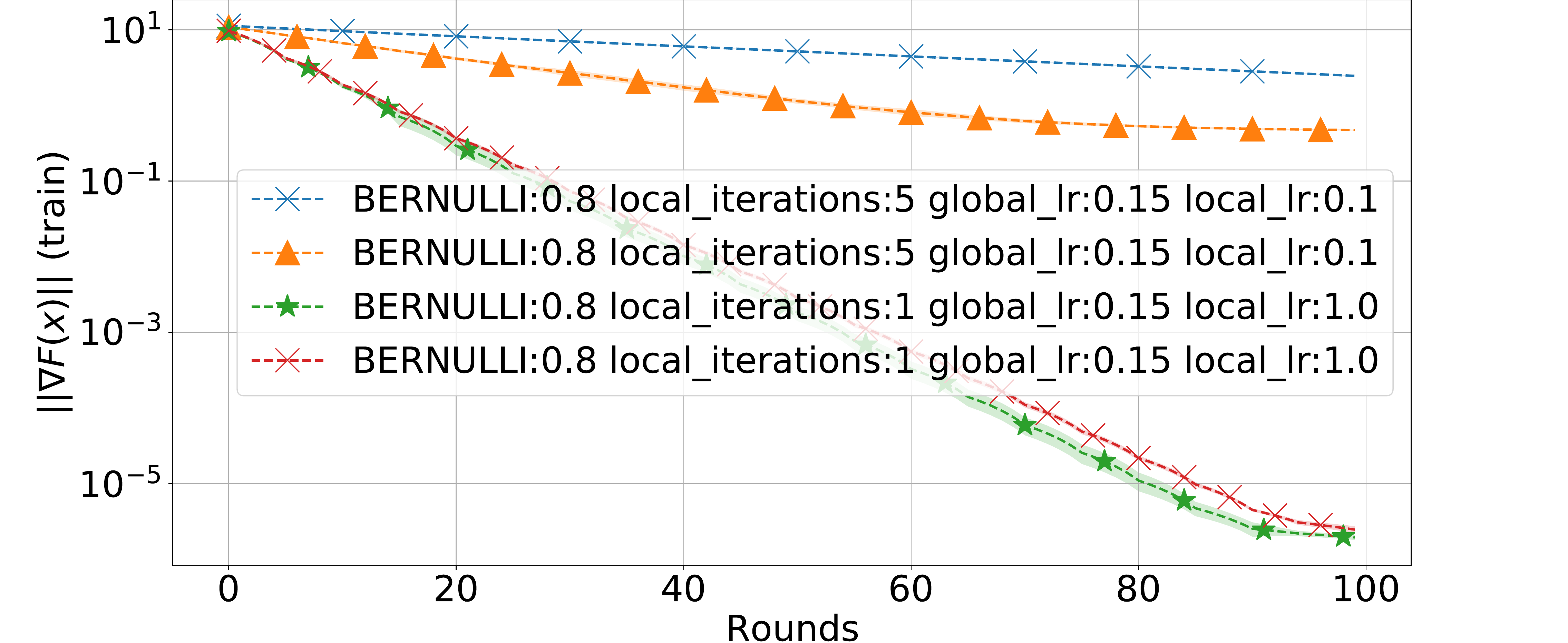} &
  \includegraphics[width=67mm]{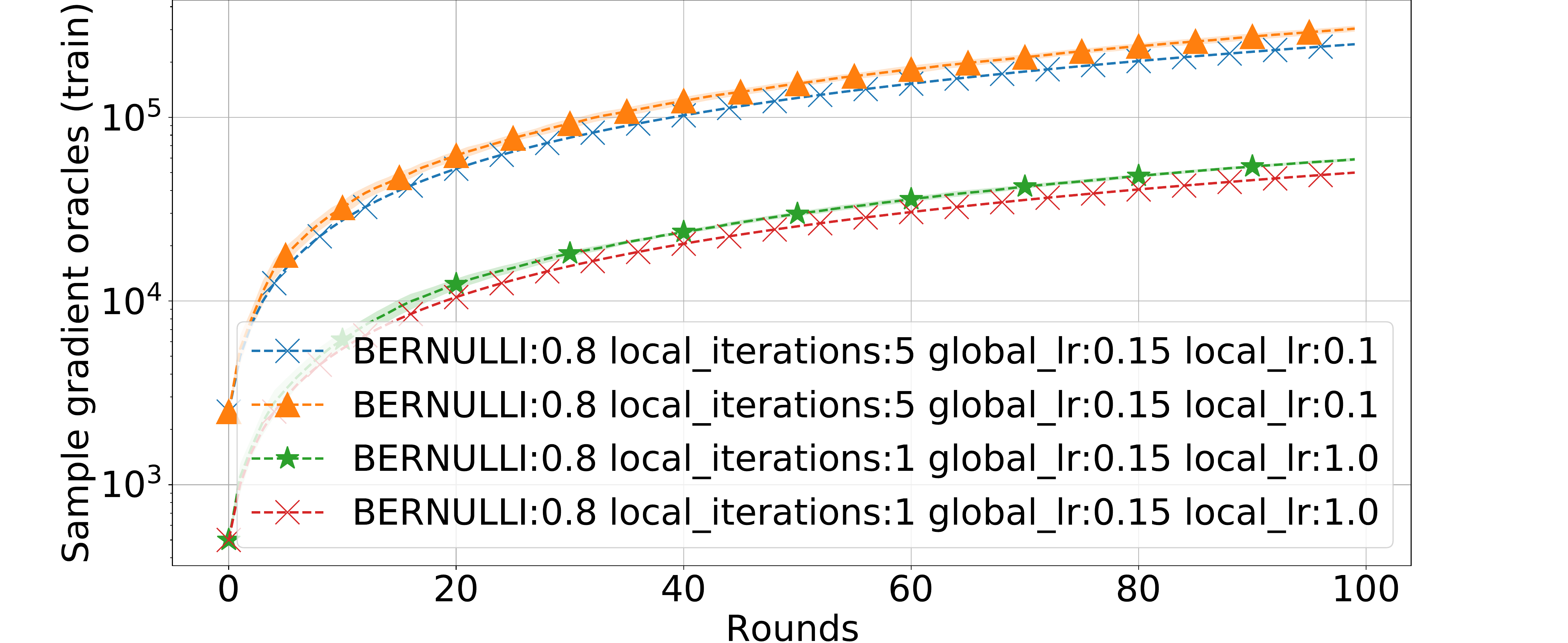} \\
(a) Convergence in gradient & 
(b) Gradient oracles \\[2pt]
\end{tabular}
\caption{Experiments with MARINA and DIANA algorithm with making local steps for quadratic minimization. Mean and variance have been estimated across 10 realizations.}
\label{fig:marina_diana_2_local_iteration_convergence}
\end{figure*}

\begin{figure*}[t]
	\begin{tabular}{cc}
		\includegraphics[width=67mm]{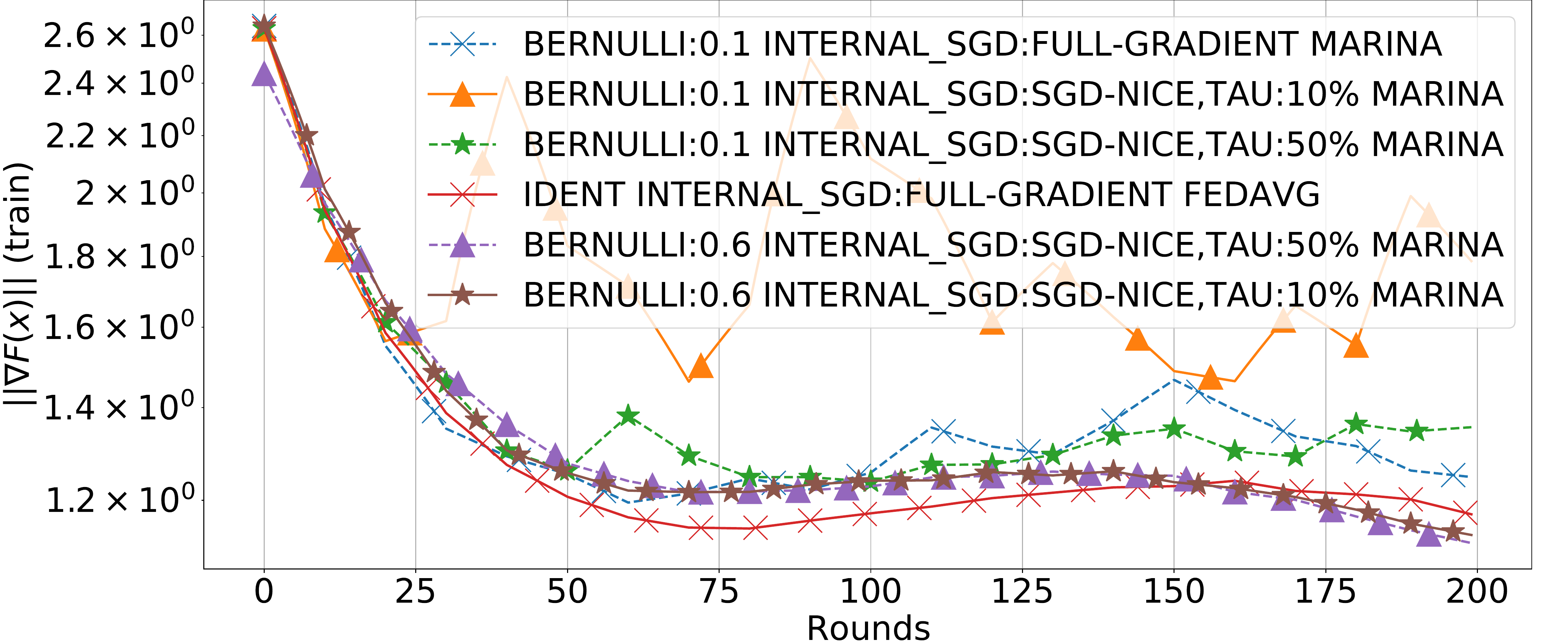} &
		\includegraphics[width=67mm]{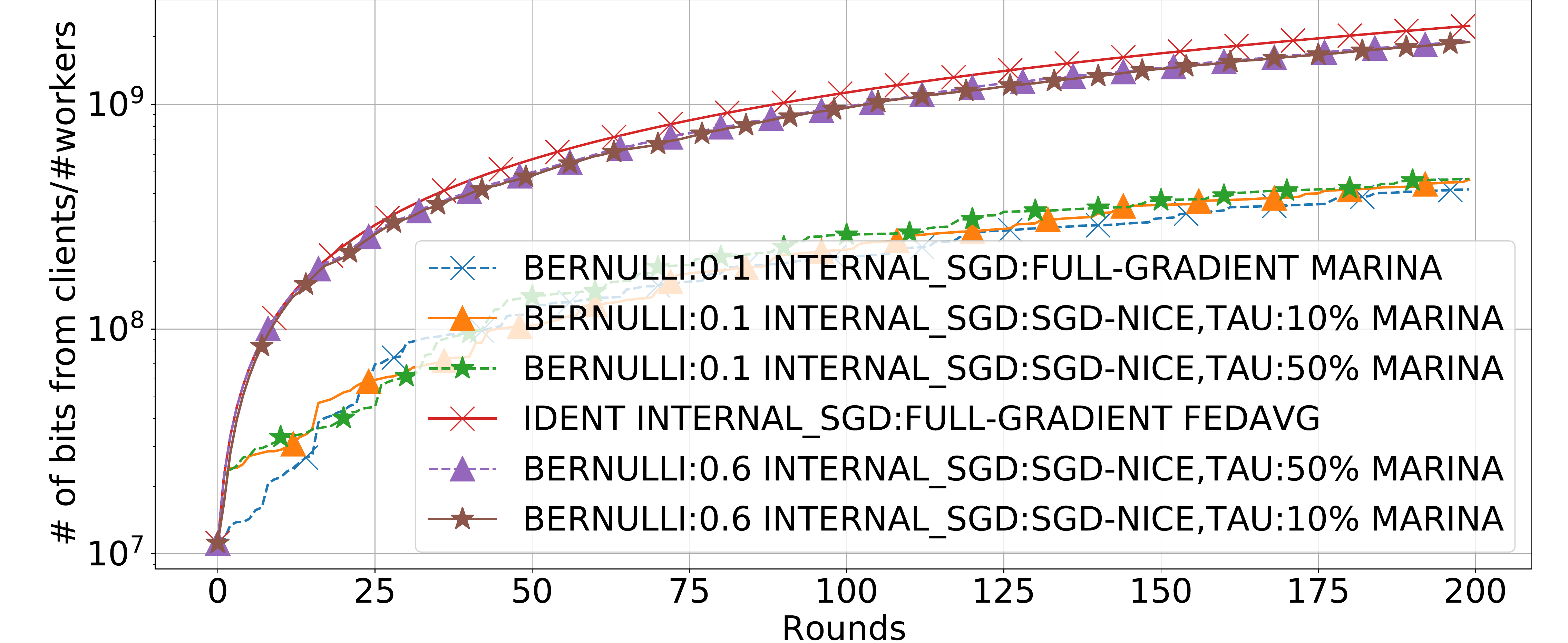} \\
		(a) Convergence in gradient & 
		(b) Average communication from clients to master \\[2pt]
	\end{tabular}
	\caption{Experiments with MARINA and DIANA algorithm with making local steps for quadratic minimization}
	\label{fig:neural_nets}
\end{figure*}
\subsection{\texttt{SCAFFOLD} with Compression}
In the first example,  we used our framework to analyze whether \texttt{SCAFFOLD} can work with compression, i.e., whether both $(\localChange_i^{(t)},U_i^{(t)})$ can be compressed, without a significant decrease in  performance.  

This experiment was carried on a synthetically generated quadratic minimization problem~\ref{eq:synthetic_quadratic}. We set the dimensionality of the problem to be $20$. Both features $a_i$'s and responses $b_i$ are generated using uniform distribution on the interval $[0, 1]$.  After this step, we update the data matrix such that the objective is L-smooth with $L = 2$ and strongly convex with $\mu=1$ using singular value decomposition (SVD). We generate $10$ clients and consider a full participation scenario. The number of communications rounds is chosen to be $100$. We set the global learning rate to $0.5$ and the local learning rate to $1.0$. For the compressor, we choose  RAND-$K$, with 3 values of $K$: $100\%$(no compression), $40\%$ and $20\%$. We provide results of this experiment in Figure~\ref{fig:scaffold_experiment_convex}.  One can observe that dropping $60\%$ coordinates at random has minimal effect on the convergence with respect to iterates.  Dropping $80\%$ brings a visible slow-down in per iterate convergence, but it is comparable with respect to the number of communicated bits. This experiment demonstrates that it may be worthwhile to consider an extension of \texttt{SCAFFOLD} by adding compression.

\subsection{Benefits of Local Updates}
For the second experiment, we consider a similar setup as for the previous experiment.  We set smoothness constant $L$ to $5$ and strong convexity parameter $\mu=1$. As a compressor we select Lazy/Bernoulli compressor described below\ref{eq:bernoulli_compressor} with $p=0.8$.
\begin{equation}
\label{eq:bernoulli_compressor}
C(x)=\begin{cases}
\nicefrac{x}{p}, &\text{with probability } p\\
0,&\text{with probability} 1 - p.
\end{cases}
\end{equation}
 We run \texttt{MARINA} and \texttt{DIANA} algorithms with full gradient estimation for $100$ rounds with $10$ clients and full participation. These methods were not analyzed in the setting with several local iterations that motivates us to consider these methods for this experiment.

Looking into Figure~\ref{fig:marina_diana_2_local_iteration_convergence}, we can observe that both \texttt{MARINA} and \texttt{DIANA} do not benefit from extra local updates. We also experimented with different problem condition numbers ($\nicefrac{L}{\mu}$), but it seems that local iterations do not speed up the convergence in the strongly convex regime. This might suggest that a naive combination of these algorithms with local updates will not lead to theoretical benefits, too.

\subsection{Stochastic Updates and Compression}
For the third example, we run \texttt{MARINA} and choose the model to be ResNet18 and for the dataset, federated version of CIFAR10 split uniformly at random (u.a.t.) among 100 clients. We run 200 communications rounds and in each round, we randomly select 25 clients. The local learning rate is $1.0$ and the global learning rate is 0.001. We perform one local iteration, i.e., $\tau_i=1$.

We investigate the effect of the combined effect of compression and stochastic gradient estimation since both of these approaches introduce extra variance that might negatively impact the quality of the obtained solution.

For gradient estimation, we employ \texttt{SGD-NICE} sampling strategy that estimates gradient via selecting uniformly at random a subset of samples of fixed size. $TAU$ in the experiments reflects the relative subset size with respect to the total size of the local dataset. The compressor operator is Bernoulli compression.  Looking into Figure~\ref{fig:neural_nets}, one can see that having Lazy compression with $p=0.1$ combined with \texttt{SGD-NICE} with $TAU=10\%$ hurts the convergence. On the other hand,  having a less aggressive compressor or using a bigger sample size for \texttt{SGD-NICE} does not deteriorate the performance.

\section{Conclusions}
In this work, we have introduced \fl, an efficient FL simulator based on \texttt{PyTorch} that enables FL researchers to experiment with optimization algorithms to advance current state-of-the-art (SOTA). \fl is an easy to use tool that supports SOTA FL algorithms and the most used image classification FL datasets. In addition, \fl uses several levels of parallelism for efficient execution while being simple to extend using only standard \texttt{PyTorch} abstractions. In its basic version, it does not require any programming and all its pre-implemented components are accessible through its GUI which allows to set up, run, monitor and evaluate different FL methods.

For the future, we plan to provide users with more pre-implemented algorithms, datasets, and models. We also aim to extend our visualization tool. After open-sourcing \fl, we expect a significant increase in the number of supported algorithms, datasets, and models via inputs from the FL research community.

\clearpage

\bibliographystyle{ACM-Reference-Format}
\bibliography{references}


\newpage
\appendix
\onecolumn

\section{\fl: Template methods of Algorithm \ref{algo:generalized_fedavg}}
\label{app:fl_skeleton}

Here we present an overview of templated methods under research responsibilities for Algorithm \ref{algo:generalized_fedavg}. For subtle technical details please familiarize yourself with the provided readme, tutorial, automatically generated code documentation in the code repository, and well-documented code.

\begin{table}[h!]
\caption{Brief description of template methods of Algorithm \ref{algo:generalized_fedavg}}
\label{tbl:generalized_fedavg_steps}
\centering
\bgroup
\definecolor{headcolor}{RGB}{47,79,79}
\def\arraystretch{1.5}%
\begin{tabular}{|p{0.24\textwidth}|p{0.76\textwidth}|}
	\hline
	{\textcolor{black}{Template Method}} &
	{\textcolor{black}{Description}} \\
	\hline
	\InitializeServerState &  This method should return a dictionary that initializes the server state. The method obtains as dimension of the problem, and the constructed and initialized model. \\
	\hline
	\ClientState &  
	   By our design client state is stateless. The client state is instantiated at the beginning of each round for each of the selected clients. You should reconstruct client state based on initialized or updated server state.
	\\
	\hline
	\LocalGradient &  
	   This method should evaluate the local gradient that is optimization algorithm specific.
	\\
	\hline
	\textsc{ClientOpt} & 
	Local classical optimizer provided out of the box by \texttt{PyTorch} used in client for local steps.\\
	\hline
	\textsc{LocalState} & The default implementation of this step in presented in some sense in \textit{local\_training} method. This step happens automatically, and in rare cases there is a need for customization this step.\\
    \hline
	\textsc{ServerOpt} & Classical optimizer provided out of the box by \texttt{PyTorch} used in sever for global steps.\\
	\hline
	\ServerGradient &  
	Server gradient is the method that estimates the direction of the global server model update, which should return a flat vector with $d$ elements.
	\\
	\hline
	\ServerGlobalState &  
	This logic is dedicated of the global server state update. This method as input obtains collected and ready to use information about clients responses and clients in that communication round, previous global model and new model with updates parameters model.
	\\
	\hline
\end{tabular}
\egroup
\end{table}

\newpage
\section{\fl: Internal Mechanisms}
\label{app:fl_vis}
In this section, we would like to depict \fl main components in a schematic way.

\begin{figure}[ht]
\centering
\includegraphics[width=0.65\textwidth, keepaspectratio=true]
{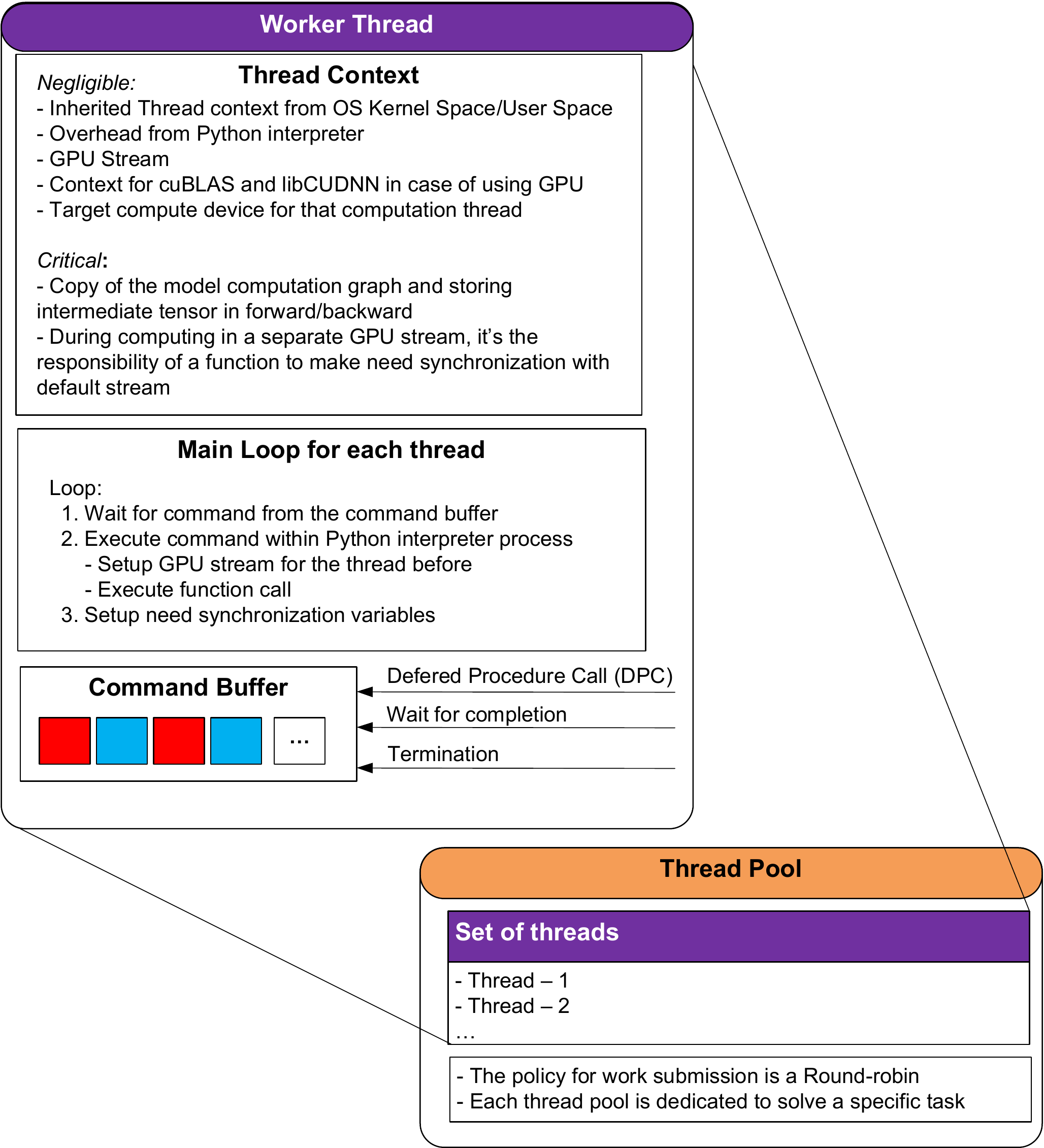}
\caption{A single worker thread structure and it's role in a thread pool.}
\label{fig:fl_worker_th}
\end{figure}

\begin{figure}[ht]
\centering
\includegraphics[width=0.4\textwidth, keepaspectratio=true]
{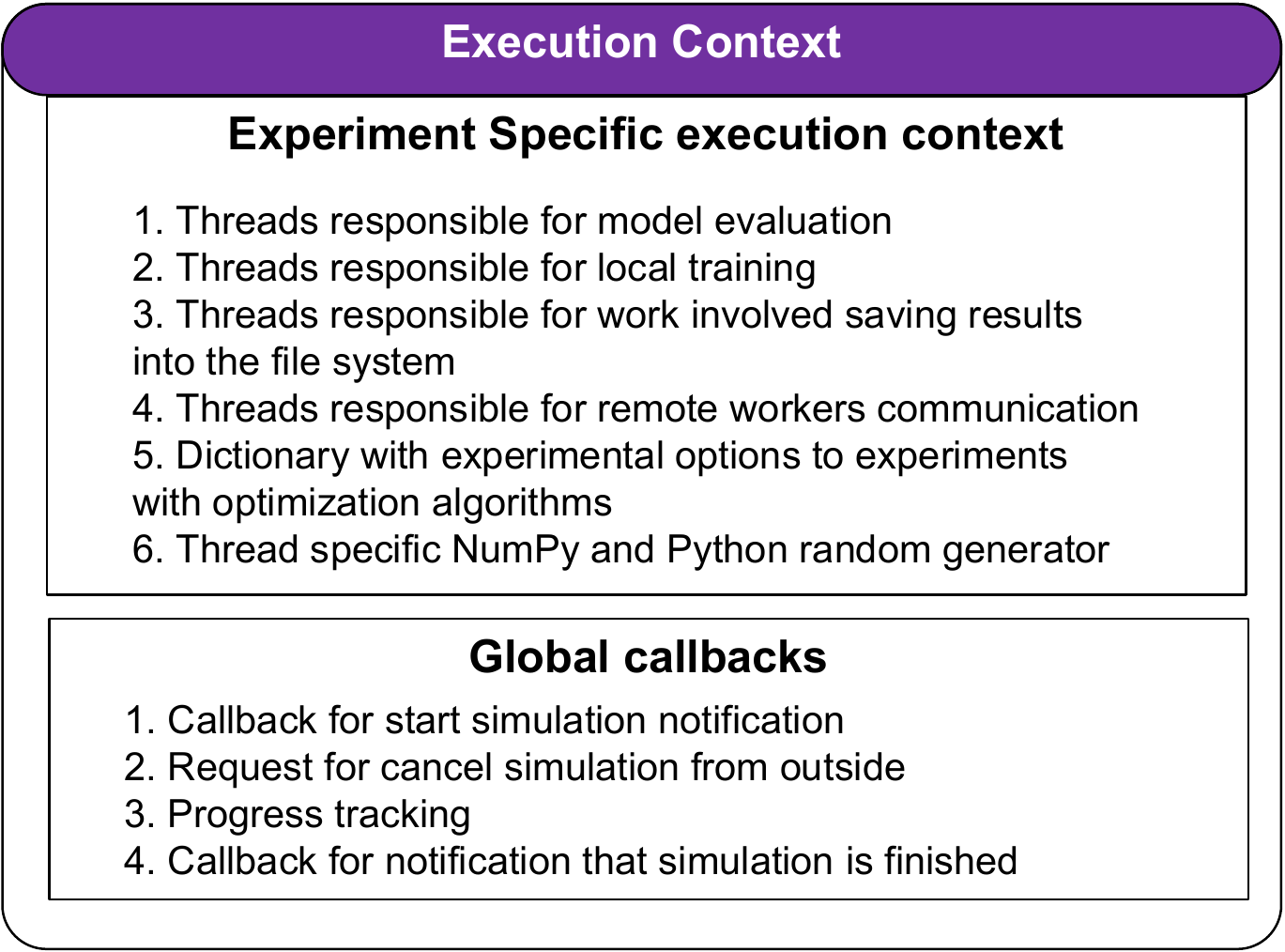}
\caption{\fl execution context for a single experiment. The GUI can handle several experiments at the same time.}
\label{fig:fl_exec_ctx}
\end{figure}

\begin{figure}[t]
\centering
\includegraphics[width=1.0\textwidth, keepaspectratio=true]
{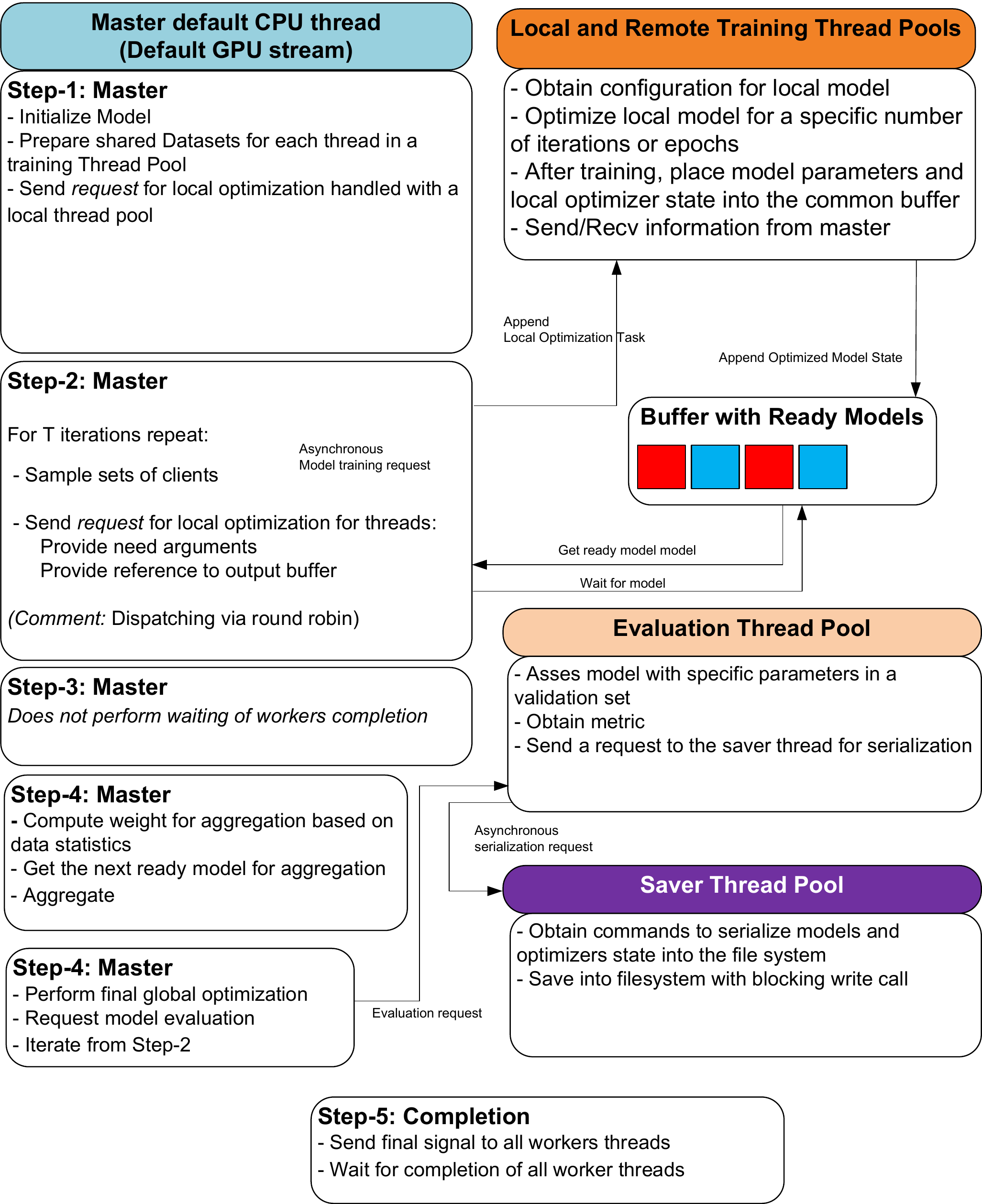}
\caption{Communication between different threads during Algorithm \ref{algo:generalized_fedavg} execution}
\label{fig:fl_components}
\end{figure}

\clearpage
\newpage

\section{Comparison to Related Frameworks}
\begin{table}[h!]
\caption{Comparison of \fl, \texttt{FedML}, and \texttt{Flower} in terms of important functionality.}
\label{tbl:cmp}
\centering
\bgroup
\definecolor{headcolor}{RGB}{47,79,79}
\def\arraystretch{1.5}%
\begin{tabular}{|p{0.05\textwidth}|p{0.35\textwidth}|p{0.15\textwidth}|p{0.15\textwidth}|p{0.15\textwidth}|}
	\hline
	{\textcolor{black}{\#}} &
	\textcolor{black}{Comparison criteria} & \textcolor{black}{\fl} & \textcolor{black}{\texttt{FedML.ai}} & \textcolor{black}{\texttt{Flower.dev}} \\
	\hline
	1 & Home page & $\dots$ & \href{https://fedml.ai/}{fedml.ai} & \href{https://flower.dev/}{flower.dev} \\
	\hline
	2 & Support of stochastic compressors & Yes & No & No \\	
	\hline
	3 & Support of local steps & Yes & Yes & Yes \\	
	\hline
	4 & Create plots w/ (w/0) internet connection & Yes (Yes) & Yes (No) & Yes (No) \\
	\hline
	5 & Create highly customizable plots & Easy & Harder & Harder \\
	\hline
	7 & Serialization of the results from numerical experiments & Yes & No & No \\
	\hline
	8 & Support of standalone and distributed mode & Yes & Yes & Yes \\
	\hline
	9 & The communication protocols for clients in multi-node setup & TCP/IP & MPI, gRPC & gRPC \\
	\hline
	10 & Synthetically controlled optimization problems & Yes & No & No \\
	\hline
	11 & Number of supported models and datasets & Modest & High & High \\
	\hline
	12 & Client paralelization in a single GPU & Yes & No & No \\
	\hline
	13 & Debugging & Easy\tablefootnote{computations can be reduced to a single thread systems setup}  & Distributed is hard &  Distributed is hard \\
	\hline
	14 & Parallelization across several GPUs (standalone) & Yes & No & No \\
	\hline
	15 & GUI interface for launch experiments & Yes\tablefootnote{PyQt5 based cross-platform GUI compatible with macOS, Linux, and Windows} & No & No \\
	\hline
	16 & Console interface for launching experiments & Yes & Yes & Yes \\
	\hline
	17 & Automatic testing & Yes\tablefootnote{Unit-tests with \textit{pytest}} & No & No\\
	\hline
	18 & Built-in mechanism to load/save experiments & Yes & No & No  \\
	\hline
\end{tabular}
\egroup
\\
\end{table}

\end{document}